\documentclass[ijds,nonblindrev]{informs4}

\OneAndAHalfSpacedXI

\usepackage{multirow}
\usepackage{amsmath,amssymb,amsfonts}
\usepackage[referable]{threeparttablex}
\usepackage{booktabs}
\usepackage{rotating}
\usepackage{fancyvrb}
\usepackage{bbm}
\usepackage{caption}
\usepackage{subcaption}
\usepackage[
natbib=true,
style=authoryear-icomp,
maxbibnames=9,
maxcitenames=2,
backend=biber
]{biblatex}
\TheoremsNumberedThrough     
\ECRepeatTheorems

\EquationsNumberedThrough    

\begin{document}



\TITLE{Forecasting Automotive Supply Chain Shortfalls with Heterogeneous Time Series}

\ARTICLEAUTHORS{%
\AUTHOR{Bach Viet Do}

\AFF{Ford GDIA, 22001 Michigan Ave Dearborn MI 48124 USA, \EMAIL{bdo1@ford.com}}

\AUTHOR{Xingyu Li}

\AFF{Ford GDIA, 22001 Michigan Ave Dearborn MI 48124 USA, \EMAIL{xli236@ford.com}}

\AUTHOR{Chaoye Pan}

\AFF{Ford GDIA, 22001 Michigan Ave Dearborn MI 48124 USA, \EMAIL{cpan9@ford.com}}
}

\ABSTRACT{%
Operational shortfalls can have a major impact on a company’s performance. Ford, with its extensive network of plants around the world, uses tens of billions of parts each year to manufacture millions of cars and trucks. With up to ten levels of suppliers between the company and raw materials, any extended disruption in this supply chain can lead to substantial financial losses. Therefore, the ability to predict and identify early shortfalls is crucial to maintaining seamless operations. In this study, we demonstrate how we constructed a dataset of multivariate time series to forecast first-tier supply chain shortfalls, using features related to capacity, variability, utilization, and processing time. This dataset is technically challenging due to its vast scale of over five hundred thousand time series. Furthermore, these time series, while exhibiting certain similarities, also display heterogeneity within specific subgroups. To address these challenges, we propose a novel methodology that integrates an enhanced Attention Sequence to Sequence Deep Learning architecture, using Neural Network Embeddings to model group effects, with a Survival Analysis model. This model is designed to learn intricate heterogeneous data patterns related to operational shortfalls. Our model has demonstrated a strong performance, achieving 0.85 precision and 0.8 recall during the Quality Assurance (QA) phase across Ford's five North American plants. Furthermore, to address the common criticism of deep learning models as ``black boxes,'' we show how the SHAP framework can be used to generate the importance of characteristics from the predictions of the proposed model. It offers valuable insights that can lead to actionable strategies and highlights the potential of advanced machine learning for managing and mitigating supply chain risks in the automotive industry. \footnote{
The information presented in this paper is intended solely for academic research purposes and should not be construed or used as financial advice.}
}


\maketitle

\section{Introduction}

Ford maintains a complex supply chain and operational network, operating 37 plants globally, utilizing more than 17 billion parts annually to manufacture about six million cars and trucks. The company has up to 10 tiers of suppliers between itself and raw materials. An extended disruption anywhere within this extensive supply chain can inflict a substantial financial damage on the company. In the literature, scholars and practitioners generally agree that disruption negatively impacts the company \citep{sheffi2005, hendricks2005, network2013building}. However, there is less agreement on classifying and forecasting such disruptions \citep{kleindorfer2005, tang2006, wagner2006, sodhi2012}.  Understanding the different sources and risks of disruption is critical because we can make informed decisions on which disruptions warrant mitigation investment \citep{simchi2014}.

A supply chain disruption refers to any incident that adversely affects the production, distribution, or sales of products. Such disruptions can stem from a wide array of sources, including global pandemics, natural disasters, geopolitical risks, terrorist attacks, environmental hazards, fluctuating fuel prices, increasing labor costs, currency fluctuations, counterfeit parts and products, delivery delays, market changes, and issues related to supplier performance. Our study focuses on forecasting the potential disruptions where the associated risks are known and controllable. Specifically, our work at Ford aims to forecast the time until occurrences where assembly parts delivery is being delayed beyond their scheduled production date at Ford assembly plants. Throughout this paper, we refer to these events as ``shortfall events.''

In the supply chain literature, efforts have been dedicated to modeling the propagation of disruption effects through supply chain networks. One of the early studies by \cite{hopp2006} analyzed the propagation of disruption effects through a simplified supply chain. \cite{schmitt2011} evaluated response strategies to minimize service level impacts in a multi-echelon network during random-duration disruptions. \cite{mackenzie2014} examined the interaction between supplier and buyer response strategies under random-duration disruptions. Simchi-Levi et al. \citep{simchi2014,simchi2015} introduced the Risk Exposure Index (REI) model to quantify the impact of disruptions at a given supplier. This index enables companies to rank their direct and indirect suppliers, identifying the "weak links" in their supply chains. Additionally, \cite{simchi2015} proposed the Time-To-Survive (TTS) concept, which defines the maximum duration the entire supply chain can generally function before the ripple effects of a disruption impact performance. Ripple effect (see \cite{ivanov2019handbook}) describes the impact of a disruption propagation on supply chain performance as well as scope of changes in the structural design and planning parameters of the supply chain. The concepts—TTR, REI, and TTS—have been implemented at Ford Motor Company to manage supply chain risks \citep{simchi2015}. More recently, \cite{dolgui2024network} proposed and analyzed a "network-of-networks" mechanism as a cross-industry adaptation strategy to mitigate uncertainties in supply chains during crises such as the COVID-19 pandemic.

In the era of Big Data, the rise of Machine Learning and Artificial Intelligence offers new opportunities to leverage massive amount of data to predict these disruptive events. Nevertheless, the literature on utilizing Machine Learning for supply chain risk management remains sparse. \cite{fan2015} explored the potential of leveraging big data sources related to supply chains and proposed a Supply Chain Risk Management (SCRM) framework to detect emerging risks. \cite{he2020} recognized the predictive capabilities of incorporating a significant data analytical component into a generic SCRM framework. Nevertheless, these works are theoretical and lack real-world application or implementation of the proposed frameworks and models. More concretely, \cite{ye2015} used a Support Vector Machine classifier to identify disruptions based on the economic performance of firms within the supply chain, collecting public financial data for these firms before, during, and after supply chain disruptions. \cite{brintrup2020} analyzed historical data from an Original Equipment Manufacturer (OEM) that produced complex engineering assets to predict order delays using machine learning models, including Random Forest, Support Vector Machine, Logistic Regression, and Linear Regression. In their study, the Random Forest Classifier achieved the best performance. Although both studies identify potential suppliers with a high risk of disruptions, they do not provide estimates for the time until these disruption events occur. Such estimates are crucial for integrating this research into the broader framework of Supply Chain Risk Management. 

As a leading automobile manufacturer, Ford can collect an extensive repository of historical data on suppliers' capacity and performance, providing a potent source for highly accurate predictive capabilities. Deep Learning has recently emerged as preeminent models in Artificial Intelligence, driven by a decade of rigorous research. Consequently, many state-of-the-art Machine Learning models are now based on Deep Learning. Our contribution in this work is to advance the research on applying Big Data to forecast supply chain shortfall events with sophisticated Deep Learning models. We demonstrate the construction of a complex dataset comprising many multivariate time series that track the arrangements between Ford and our first-tier suppliers for transporting critical vehicle parts to Ford's manufacturing plants. We meticulously select data features that reflect capacity, inventory, utilization, and process time, key aspects in the classical Factory Physics (see \cite{hopp2011factory}). Furthermore, we propose an AI model that integrates an enhanced Sequence to Sequence with Attention Deep Learning architecture with a parametric survival analysis likelihood model. This enhanced architecture is designed to model the heterogeneity in the data, driven by various combinations of plants, sites, and vehicle parts. Our work introduces an important improvement to the Seq2Surv model proposed by \cite{li2022}, developed for generic time series that exhibit similar underlying statistical patterns despite random variations. In contrast, the shortfall behaviors in our data can differ significantly, even contradictorily, based on the specific combinations of suppliers, plants, and parts.  While originally developed for supply chain time series data, our methodology can be generalized to any dataset comprising multivariate time series with inherent heterogeneity. Lastly, we illustrate how to use the SHAP (SHapley Additive exPlanations, \cite{phillips2021}) framework to calculate feature importance for each of the model’s predictions. This explainable AI technique provides valuable insights that can inform actionable strategies for our business partners.

Our model relies on the Survival Analysis framework to model observed time-to-disruption data. Survival Analysis, a branch of Statistics, focuses on the study of time-to-event data \citep{jenkins2005}. This field encompasses a variety of applications, such as predicting the survival of cancer patients \citep{vigano2000}, customer churn \citep{van2004}, mechanical system failure \citep{susto2014}, credit scoring \citep{dirick2017}, and reliability and manufacturing problems \citep{krivtsov2002, li2022}. The strength of Survival Analysis lies in its interpretability, flexibility, and ability to handle censored data. However, one of its notable weaknesses is predictive accuracy. To enhance predictive performance, numerous studies have extended classical Survival Analysis using Neural Networks. 

\cite{faraggi1995} were among the pioneers, extending Cox regression by replacing its linear predictor with a single hidden layer multilayer perceptron (MLP). \cite{katzman2018} revisited this approach within the deep learning framework, introducing DeepSurv, a model that outperformed traditional Cox models in terms of the C-index (see \cite{harrell1982}). Other similar works include SurvivalNet by \cite{yousefi2017} , which fits Cox proportional models using Neural Networks and applies Bayesian optimization on tuning hyperparameters. \cite{zhu2016, zhu2017} utilized Convolutional Neural Networks instead of MLP in their work. \cite{kvamme2019} proposed an extension of the Cox model where the proportionality constraint is relaxed, introducing an alternative loss function that scales well for both proportional and non-proportional cases. \cite{li2022} further extended these methodologies by leveraging Sequence-to-Sequence with Attention deep learning architecture for time series survival analysis data.

For discrete-time survival problems, \cite{lee2018} applied neural networks to the discrete-time likelihood for right-censored time-to-event data, parameterizing the probability mass function. \cite{gensheimer2019} adopted a similar likelihood approach, parameterizing hazard rates with a neural network. \cite{kvamme2019discrete}, through simulation studies and real-world data, found that hazard rate parameterization performed slightly better. Building on this insight, the authors introduced PC-Hazard, in which the parameterization of the hazard rate for continuous survival time data is achieved by discretizing the continuous time scale, assuming the continuous-time hazard is piecewise constant.

The remainder of the paper is organized as follows: Section 2 reviews the relevant background and preliminaries. Section 3 offers a detailed explanation of how to construct a dataset for forecasting supply chain disruptions by selecting informative features. In Section 4, we present the Heterogeneous Sequence-to-Disruption AI model. Section 5 evaluates the model's performance and discusses methods for interpreting the results using the SHAP framework. Finally, Section 6 concludes the paper.

\section{Background} \label{prelim}
\subsection{Survival Analysis} \label{survival}

In this study, we aim to model the distribution of the time-to-shortfall, denoted as $T^*$. In practical scenarios, not all shortfall times are observable and are often subject to right censoring, wherein the observation period concludes before the shortfall event occurs. Let $C^*$ represents the time to the censoring event. Formally, we define:

\[ 
\begin{aligned} T &= \min(T^*, C^*) \\ 
Y &= \mathbbm{1}(T^* \leq C^*).
\end{aligned} 
\]

Here, $T$ is the observed time-to-shortfall, and $Y$ is an indicator that equals $1$ if the shortfall event occurs within the observation window $T^* \leq C^*$ and $0$ otherwise. This framework is fundamental in Survival Analysis. As demonstrated in the next section, in our data, the time-to-shortfall is discrete, ranging from $1$ to $365$ days. The discrete disruption time can be modeled using a Survival Analysis parametric approach (see Chapter 10 in \cite{moore2016}).

Before we define the model, we review some preliminaries. Consider a random discrete variable  $T^*$ with the discrete support $\tau_1 < \tau_2 <  \ldots < \tau_j < \ldots$, the cumulative distribution function (CDF) is given by $F(t) = P(T^* \le t) = \sum_{i=0}^t \mathbb{P}(T^* = i)$, and the corresponding probability mass function (PMF) is $f(t) = \mathbb{P}(T^* = t)$ where $t$ is a positive integer.

In Survival Analysis, it's often more useful to work with the survival function, defined as $S(t) = 1 - F(t) = P(T^* > t)$, along with its associated hazard function. For a discrete variable $T^*$ and two consecutive survival time points $\tau_{j-1} < \tau_j$, the hazard function is denoted $h(\tau_j) = \mathbb{P}(T^* = \tau_j \; | \; T^* > \tau_{j-1}) = \cfrac{f(\tau_j)}{S(\tau_{j-1})} = \cfrac{S(\tau_{j-1}) - S(\tau_j)}{S(\tau_{j-1})}$. As such,

\begin{equation} \label{eqn:pmf}
f(\tau_j) = h(\tau_j) S(\tau_{j-1})
\end{equation}
\begin{equation} \label{eqn:surv}
S(\tau_j) = [1 - h(\tau_j)]S(\tau_{j-1}).
\end{equation}

On the other hand, we see that,
\[ 
\begin{aligned}
\mathbb{P}(T = t, Y = y) &= \mathbb{P}(T^* = t, t \le C^*)^y \mathbb{P}(T^* > t, C^* = t)^{1 - y} \\
&= f(t)^y \mathbb{P}(C^* \ge t)^y \mathbb{P}(T^* > t, C^* = t)^{1 - y} \\
&= f(t)^y \mathbb{P}(C^* \ge t)^y \mathbb{P}(T^* > t)^{1 - y} \mathbb{P}(C^* = t)^{1 - y} \\
&= f(t)^y S(t)^{1 - y} \mathbb{P}(C^* \ge t)^y \mathbb{P}(C^* = t)^{1 - y}.
\end{aligned}
\]

The first equality follows the definition of $T^*$ and $C^*$ under the events $\{Y = 1\}$ and $\{ Y = 0\}$. The second equality is due to the independence of $T^*$ and $C^*$. The third equality also holds due to their independence. Because $T$ is discrete, we can model its hazard function using a parametric distribution with parameter $\theta$. We define $\kappa(t)$ as the index of the discrete time point in the support, i.e., $t = \tau_{\kappa(t)}$. For a sample of observations $(t_1, y_1), (t_2, y_2), \ldots, (t_n, y_n)$, the log-likelihood function can be expressed as a sum of individual log-likelihood contributions as follows using equations (\ref{eqn:pmf}) and (\ref{eqn:surv}).

\[
\begin{aligned}
l(\theta) &= \sum_{i=1}^{n} y_i \log f_\theta(t_i) + \sum_{i=1}^n (1-y_i) \log S_\theta(t_i) + \sum_{i=1}^n \left[y_i \log P(C^*_i \ge t_i) + (1 - y_i) \log \mathbb{P}(C^*_i = t_i) \right] \\
&\propto \sum_{i=1}^{n} y_i \log f_\theta(t_i) + \sum_{i=1}^n (1-y_i) \log S_\theta(t_i) \\
&=  \sum_{i=1}^{n} y_i \log h_\theta(t_i) + y_i \log S(\tau_{\kappa(t_i) - 1}) + \sum_{i=1}^n (1-y_i) \log[1 - h_\theta(t_i)] +  (1 - y_i)\log S_\theta(\tau_{\kappa(t_i) - 1}) \\
&= \sum_{i=1}^n \left[ y_i \log h_{\theta}(t_i) + (1 - y_i) \log[1 - h_{\theta}(t_i)] + S(t_{\kappa(t_i) - 1}) \right].
\end{aligned}
\]

In the second line, terms related to censoring times $C_i$ are omitted from the log-likelihood function as they are constant with respect to the parameter $\theta$. Utilizing equation \ref{eqn:surv} recursively in the final term of the last equality above, we can finally write the log-likelihood of the sample data as,

\begin{equation} \label{eqn:likelihood}
l(\theta) = \sum_{i=1}^n \left[ y_i \log h_{\theta}(t_i) + (1 - y_i) \log[1 - h_{\theta}(t_i)] + \sum_{j=1}^{\kappa(t_i) - 1} \log[1 - h_{\theta}(\tau_j)]\right].
\end{equation}

Observe that the equation $\ref{eqn:likelihood}$ consists solely of the hazard function. To model the observed data, we simply need to define the form of the hazard function. Since this hazard function represents a probability of discrete variable $T$, its range must lie between 0 and 1. One suitable modeling choice is the logistic hazard function (see \cite{brown1975}). Assume that the time-to-shortfall is also conditioned on covariates/features $X$,

\begin{equation} \label{eqn:hazardfn}
h(\tau_k) = \cfrac{1}{1 + \exp\left( - \theta_k^T X \right)}, k = 1, 2, \ldots,
\end{equation}

where $\theta$ is the model parameter and the linear combination $\theta^T X$ reflects the linear relationship between the features and the model parameter.

\subsection{Sequence-to-Sequence Architecture}

The Seq2Seq model, introduced by \cite{sutskever2014} at Google, represents a pivotal milestone in deep learning. This model processes a sequence of inputs to generate a corresponding sequence of outputs. While initially developed to address the challenges of machine translation, the Seq2Seq model has since become a foundational framework for various natural language processing tasks and has been adopted for time series data in other fields.

The core architecture of the Seq2Seq model is the encoder-decoder framework, which is frequently implemented using Recurrent Neural Networks (RNNs, see \cite{rumelhart1986}). The encoder's primary function is to convert the sequential input data into hidden states and a context vector. The context vector, the aggregated sum of these hidden states, encapsulates the information from the input sequence into a fixed-size representation.

Once the input sequence has been encoded, the decoder generates the output sequence. It leverages the context vector and hidden states to produce the output sequence. Operating in an autoregressive manner, the decoder generates one unit of the output sequence at a time. This step-by-step generation ensures that each subsequent unit is conditioned on the previously generated units, thereby maintaining coherence in the output sequence.

A significant enhancement to the Seq2Seq model is the integration of Attention mechanism proposed by \cite{bahdanau2014}. This mechanism enables the model to dynamically focus on different parts of the input sequence while generating each output unit. Specifically, the context vector is computed as a weighted sum of the encoder's hidden states, with the weights reflecting the relevance of each hidden state to the current decoding step. This dynamic focusing capability significantly improves the model's performance, particularly for tasks involving long and complex input sequences.

\subsection{Seq2Surv Model} \label{seq2surv}
Leveraging the power of Sequence-to-Sequence neural networks and the Attention mechanism for modeling sequential data, \cite{li2022} employed this advanced deep learning architecture to handle time series data in the reliability and manufacturing domain. This approach replaces the linear predictor in Cox proportional hazard models (see Chapter 5 in \cite{moore2016}) and discrete time logistic hazard parametric models (see equation \ref{eqn:hazardfn}) with learnable non-linear functions. In this methodology, time series are treated as sequential input to the Encoder, which \cite{li2022} implemented using Bidirectional Gated Recurrent Units (GRUs), as introduced by \cite{cho2014}. The GRU, similar to a Long Short-Term Memory network (LSTM, see \cite{hochreiter1997}) with its gating mechanisms for input and forgetting features, lacks output gates, resulting in fewer parameters than LSTMs. The Decoder in Seq2Surv then generates a sequence of survival probability estimates for the entire lifetime, which is then used in the Survival Analysis' log-likelihood function to model the observed disruption times.

\section{Data Description and Feature Engineering} \label{data}

\begin{table}[!ht]
  \centering
  \resizebox{\columnwidth}{!}{%
  \begin{threeparttable}
    \caption{Full List of predictive features and their brief descriptions. The features are selected the reflect of aspects of Inventory, Capacity, Utilization and Processing Time.}
    \label{tab:feats}
    \begin{tabular}{cl}
      \toprule
      \textbf{Feature Name} & \textbf{Description} \\
      \midrule
      \textbf{APPC} & Average Purchased Part Capacity refers to the production capacity for a part that a supplier can provide to Ford during a normal working week, as reported by the suppliers. \\
      \textbf{APW} & Average Production Weekly for a part, the estimated weekly production demand during normal working days, as determined by Ford. \\
      \textbf{MPPC} & Maximum Purchased Part Capacity, the maximum weekly capacity for a part that a supplier can provide to Ford, as reported by the supplier. \\
      \textbf{MPW} & Maximum Production Weekly, the estimated maximum capacity demand for a part for production, as determined by Ford. \\
      \textbf{cum\_release} & Cumulative Released Quantity, this number, calculated from the beginning of the year to the present, represents Ford's demand for the supplier to ship this quantity of a part to a Ford plant. \\
      \textbf{days\_behind\_release} & Days-Behind-Release, the number of consecutive days, up to the present, that a supplier has failed to ship the required parts as requested by Ford.  \\
      \textbf{production\_usage} & Production Usage, the actual daily quantity of a vehicle part utilized in Ford's production, \\
      \textbf{qt\_behind\_release} & Quantity-Behind-Release, The quantity of a part that a supplier has failed to ship to a Ford plant as requested. \\
      \textbf{qt\_boh\_arrived} & Quantity Balance-On-Hand Release, the quantity of a part currently present on the Ford plant's campus, encompassing both unloaded items and those still loaded on transportation trucks. \\
      \textbf{qt\_prt\_boh\_loose} & The quantity of a part that is present at Ford's campus, unloaded from transporting trucks, and ready for production. \\
      \textbf{qt\_boh\_proj} & Quantity Balance-On-Hand Projection, the inventory quantity of a part, including those already available at the plant campus and those in transit but expected to arrive on time.\\
      \textbf{qt\_boh\_warehouses} & Quantity Balance-On-Hand Warehouses, the inventory quantity of a part available at Ford's warehouses, but not present at the plant campuses.  \\
      \textbf{qt\_boh\_days\_arriv} & Quantity Balance-On-Hand Arrived, the number of production days that can be protected or sustained with the part's quantity value of qt\_boh\_arrived.\\
      \textbf{qt\_boh\_days\_loose} & the number of production days Ford can sustain with the quantity of parts in qt\_prt\_boh\_loose that are available at the plant campus and have been unloaded from transporting trucks. \\
      \textbf{qt\_boh\_days\_proj} & the number of production days Ford can sustain with the available part quantities at the plant campus, combined with those still in transit but expected to arrive on time for production.\\
      \textbf{qt\_p\_y\_in\_transit} & the quantity of a part that was shipped out last year but is still in transit in the current year.\\
      \textbf{qt\_promised} & Quantity Promised, the daily quantity of a part that a supplier has promised and committed to ship to a Ford plant. \\
      \textbf{qt\_release} & Quantity Release, the daily release quantity of a part required by Ford for shipment from a supplier.\\
      \textbf{qt\_supp\_cum\_rcpt} & The cumulative quantity of parts received from a supplier to a Ford plant from the beginning of the year to the present. \\
      \textbf{qt\_supp\_in\_transit} & The part's quantity currently in transit from a supplier to a Ford plant as of today. \\
      \textbf{qt\_trans\_days\_used} & Quantity-Transit-Days-Used, the estimated number of days required for transportation from a supplier to deliver the part to a Ford plant. \\
      \bottomrule
    \end{tabular}
  \end{threeparttable}
}
\end{table}

In this section, we first describe how we construct and gather the data necessary for tracking suppliers' delivery performance and eventually forecasting time-to-shortfall events. Ford contracts with many suppliers to ship critical vehicle parts to our manufacturing plants. Each unique part ID is assigned a ``will-make'' date, indicating the scheduled date for its use in the vehicle assembly line. Shortfall events occur when a part's promised delivery date exceeds its designated will-make date.

\begin{figure}
    \centering
    \includegraphics[scale=0.31]{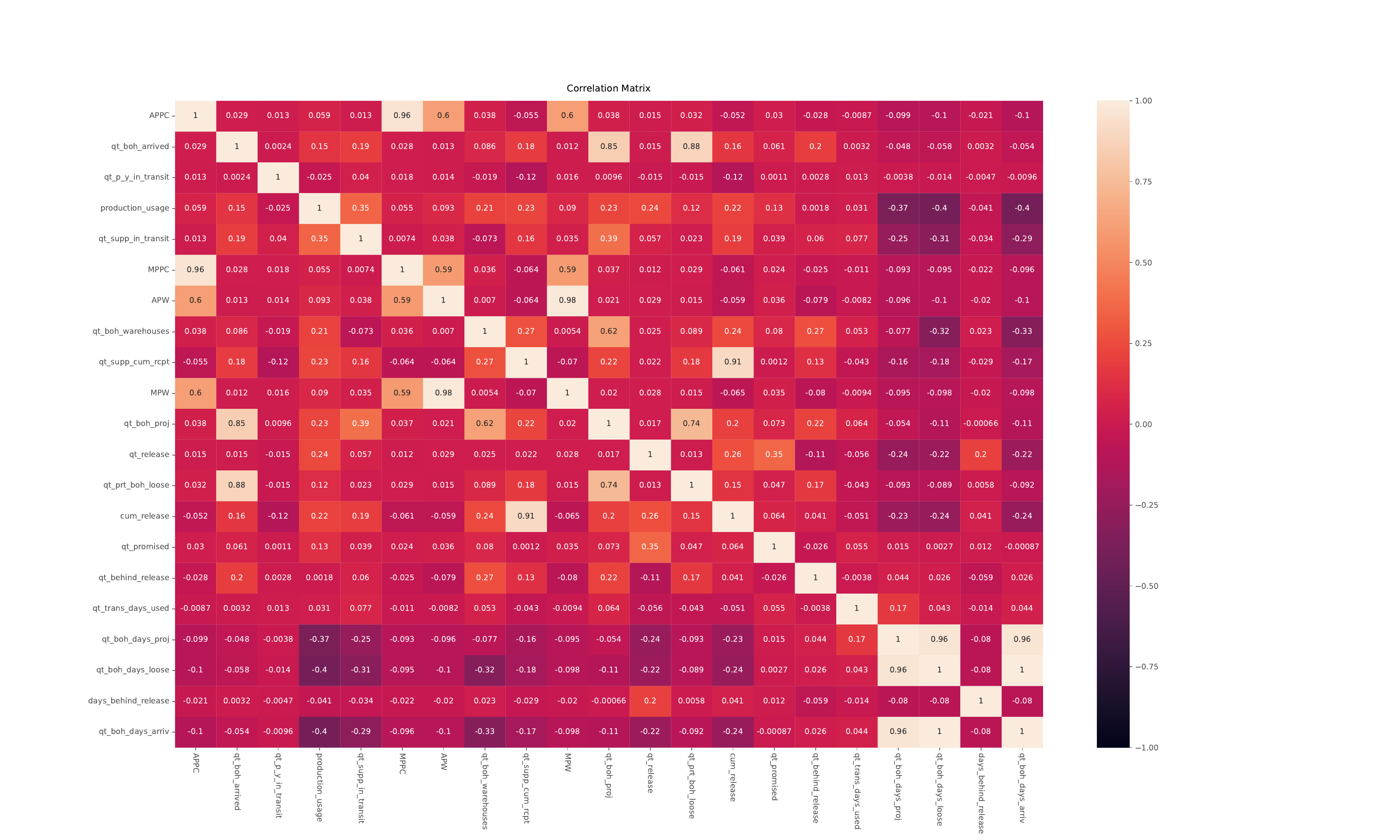}
    \caption{The Pearson correlation coefficients were computed for the $21$ features selected for the predictive model proposed in Section \ref{model}. These features were meticulously chosen to represent key aspects of the classical Physics Factory, including Inventory, Capacity, Utilization, and Processing Time.}
    \label{fig:feat_corr}
\end{figure}

A pivotal result in classical manufacturing management is the understanding that, in the presence of variability, three key factors are essential for synchronizing demand and production at lower costs and higher service levels: capacity, inventory, and cycle time \citep{hopp2011factory}. This concept is technically articulated through Little's Law and Kingman's equation (see \cite{hopp2011factory}). According to Little's Law, inventory ($L$), measured in quantity units, is proportional to throughput ($\lambda$), in quantity units per time, and cycle time ($W$): $L = \lambda \cdot W$. Kingman's equation, also known as the VUT equation, provides the expected waiting time or cycle time ($W$) as a function of variability, utilization, and process time: $\mathbb{E}[W] = V \cdot U \cdot T$ (as discussed in \cite{hopp2011factory}). This equation was developed by the British mathematician John Kingman in $1961$ using queuing theory (see \cite{kingman1961}). The first factor, variability, captures the randomness of arrival and service times; the second factor, utilization, reflects the usage of the workstation or assembly line; and the third factor, process time, represents the average processing time of the assembly line.

In practice, accurately estimating variability, capacity, and utilization can be difficult. To address this, we use Statistical Machine Learning to build a model that learns from historical trends and patterns to predict shortfalls, as discussed in Section 4. For the model to be effective, the training dataset is designed with data features strongly correlated with variability, capacity, utilization, and processing time. It comprises time series collected from January $1$, $2021$, to December $31$, $2023$. Each time series corresponds to a combination of supplier, Ford plant, and vehicle part IDs, representing an agreement between a supplier and a Ford plant to deliver vehicle parts. Table \ref{tab:feats} lists the complete features used in this study. We also track the time-to-shortfall at each time point, with the exact time in days until a shortfall event. The time-to-events are known since they are historical events.

To prepare our dataset for model training, we segment each original time series into a collection of $28$-day sequences. Within these sequences, the 21 features for the $28$ days are concatenated into a single row in a tabular dataset. Each resulting $28 \cdot 21 = 588$ feature is normalized to a range between $0$ and $1$ by subtracting the minimum value and dividing by the range between the maximum and minimum values.

For each row feature, the response variable is the time in days until a shortfall event from the last day of the $28$-day sequence— these times until shortfalls are truncated to lie between 0 and 366 days. The value 366 serves as a unique token indicating that the time series is predicted not to experience any shortfalls within an annual fiscal year.

\begin{figure}
    \centering
    \includegraphics[scale=0.6]{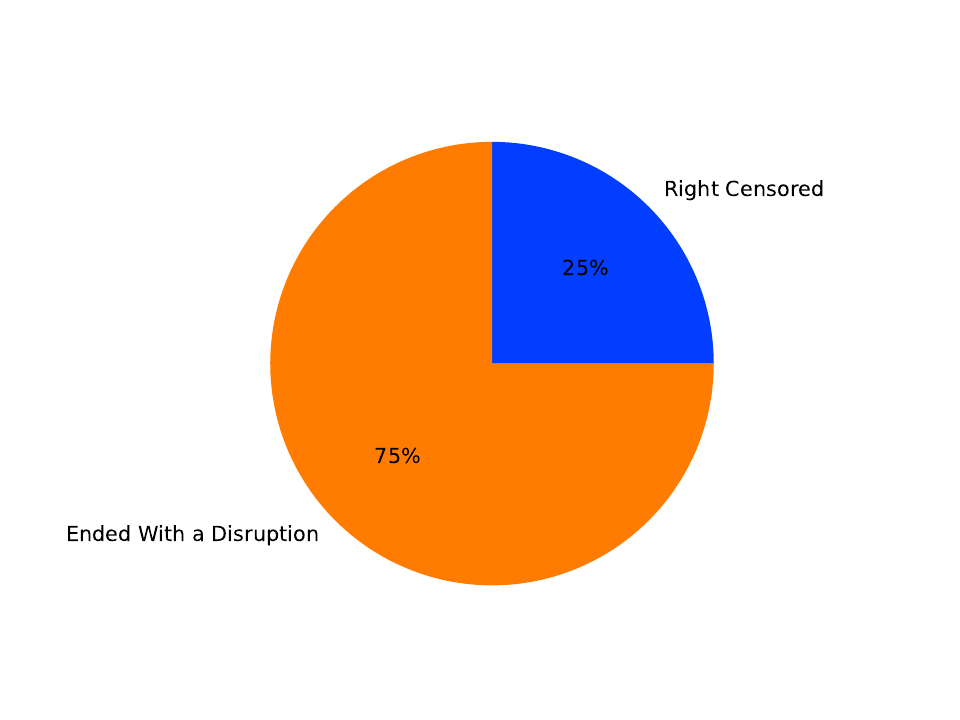}
    \caption{Proportion of Time Series: Right-Censored vs. Terminated by Disruption Events ($25$\% Right-Censored)}
    \label{fig:event_prop}
\end{figure}

Notationally, our data consists of $(d_i, X_i, t_i, y_i), \; i = 1, 2, \ldots, n$, where $X_i$ represents the $28$-day sequence features/covariates, $t_i$ is the time in days until a shortfall, $y_i$ is an indicator with one signifying that the time series resulted in a shortfall and 0 otherwise, and $d_i$ is a tuple containing three identifiers for the site, Ford plant, and vehicle part.  Figure \ref{fig:feat_corr} illustrates the Pearson correlation among $21$ features, revealing no pairs of features with perfect correlation. Furthermore, Figure \ref{fig:event_prop} demonstrates that roughly $25$\% of all the time series are right-censored.

\section{Methodology} \label{model}

This section details our methodology for modeling the dataset discussed previously, to predict the time until shortfall.

Given a sample $(d_1, X_1, t_1, y_1), (d_2, X_2, t_2, y_2), \ldots, (d_n, X_n, t_n, y_n)$, we model the times until disruption using a Survival Analysis model, as described in Section \ref{survival}. Specifically, our chosen hazard function has a logistic form but with a non-linear term $\phi_k$. The function $\phi_k(\cdot), k = 1, 2, \ldots, 366$ is a smooth, nonlinear function that models the relationship between the feature $X_i$ and the time until disruption $t_i$.

\begin{equation} \label{eqn:seq2survhazard}
h(\tau_k | X_i) = \frac{1}{1 + \exp(-\phi_k(X_i))}, \; k = 1, 2, \ldots, 366    
\end{equation}

We estimate the functions $\phi_1, \ldots, \phi_{366}$ by a Deep Learning Neural Network, Seq2Seq with bidirectional GRUs and Attention mechanism described in Section \ref{seq2surv}. The feature $X_i$ represents a sequence of 21 metrics collected over 28 days, as detailed in Section \ref{data}. This sequence is fed into the Encoder, which transforms it into 28 hidden states and a context vector—a weighted average of these hidden states with Attention Mechanism. The Encoder then processes these hidden states and the context vector to generate 366 elements, $\phi_1, \phi_2, \ldots, \phi_{366}$. These values, $\{ \phi_k \}_{k=1}^{366}$, are used to compute the hazard functions, which are crucial to the final loss function employed to train the neural network via backpropagation. The loss function is the negative log-likelihood (\ref{eqn:likelihood}), given by:

\[
\text{loss function}(\theta) = -\sum_{i=1}^n \left[ y_i \log h_{\theta}(t_i) + (1 - y_i) \log[1 - h_{\theta}(t_i)] + \sum_{j=1}^{\kappa(t_i) - 1} \log[1 - h_{\theta}(\tau_j)]\right] 
\]

In terms of model architecture, we propose a important change to the Seq2Surv model \citep{li2022}, specifically enhancing the Encoder Network to better model heterogeneity among plants, sites, and parts. By incorporating Neural Embeddings for plant, site, and part IDs, we aim to effectively capture the dataset's diversity. We refer to our modified model as \textbf{Heterogeneous Sequence-to-Survival}.

\begin{figure}[ht!]
    \centering
    \includegraphics[scale=0.2]{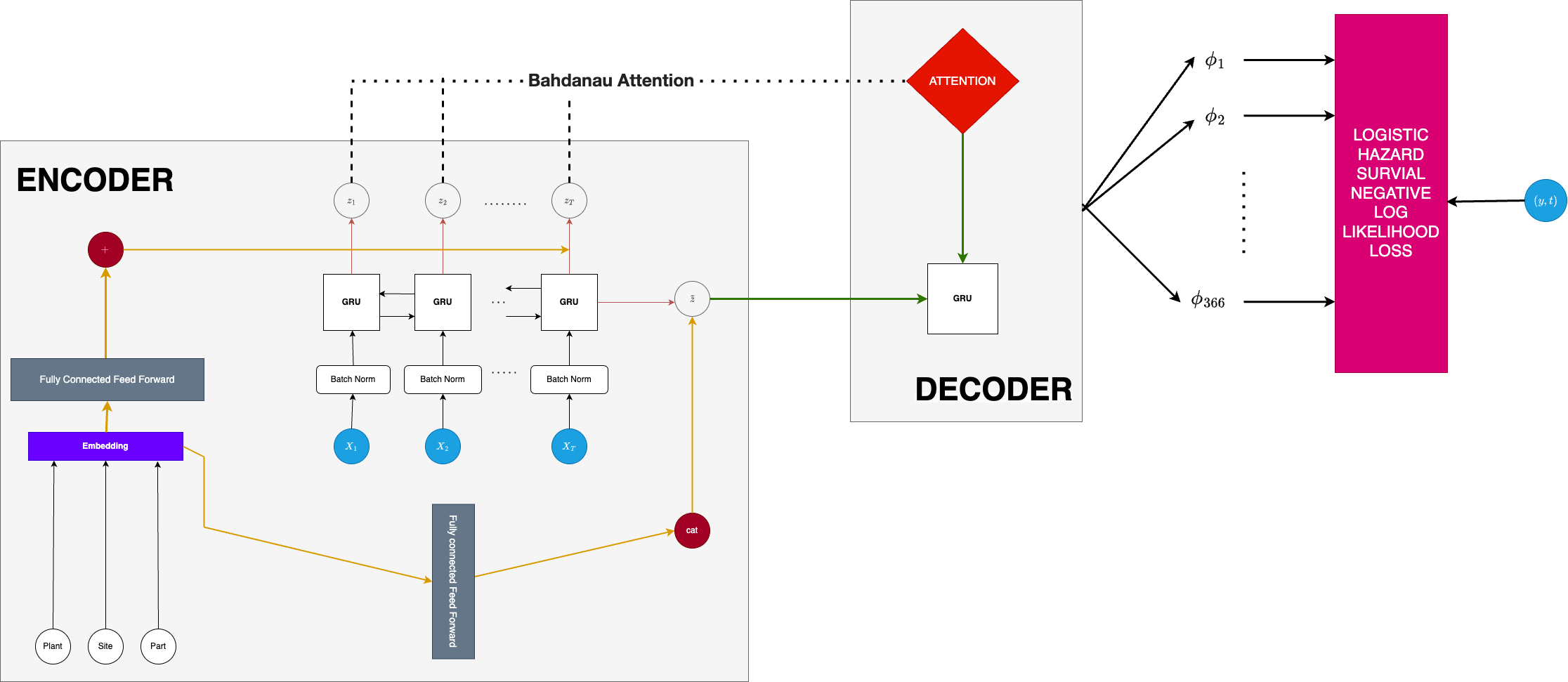}
    \caption{Heterogeneous Sequence-to-Survival comprises a Sequence-to-Sequence Neural Network augmented with embedding layers for plants, sites, and parts to account for group effects. The sequential outputs, $\phi_1, \ldots, \phi_{366}$, are subsequently input into the Parametric Survival Model, where the hazard functions are parameterized using logistic functions as in Equation \ref{eqn:seq2survhazard}.}
    \label{fig:model}
\end{figure}

As illustrated in Figure \ref{fig:model} each site ID, plant ID, and vehicle part ID is transformed into vector embeddings using neural network embedding layers. These embeddings are then concatenated and passed through two fully connected neural network layers, which generate non-linear group effect signals. These signals are subsequently added to the hidden states of the bi-directional GRUs and concatenated with the context vector. This methodological adjustment is crucial for accounting for the heterogeneity in the dataset. Empirical observations indicate that the disruption times for different combinations of sites, plants, and parts can exhibit distinct and sometimes contradictory behaviors, highlighting the need for our approach to capture these nuanced variations. The Encoder encapsulates the essential representation for each $28$-day sequence feature $X_i$. Intuitively, this representation consists of two components: an individual-specific component and a group component related to the combination of site, plant, and part values. The individual-specific information is directly inferred from the feature values of $X_i$, ensuring that the unique characteristics of each sequence are accurately modeled. Meanwhile, the group component is derived from the site, plant, and part values, allowing the model to incorporate the distinct group behaviors from the data. The individual representation captures the idiosyncrasies of each sequence, while the group representation integrates the broader, systematic effects associated with different combinations of site, plant, and part values. This modified architecture enhances the model's ability to predict shortfall times with greater accuracy and reliability, accommodating the inherent variability within the dataset.

\section{Model Performance \& Explainability}

\subsection{Predictive Performance}

\begin{table}[!ht]
\centering
\resizebox{\columnwidth}{!}{%
\begin{tabular}{|c||c|c|c|c|c|c|c|} 
\hline
\multirow{2}{*}{\textbf{Ford Plant}} & \multirow{2}{*}{\textbf{Precision}} & \multirow{ 2}{*}{\textbf{Recall}} & \multicolumn{4}{|c|}{\textbf{Normalized Confusion Matrix}}  \\
\cline{4-7}
& &  & True Positive & False Negative & False Positive & True Negative \\
 \hline
  \textit{North American Plant A} & 0.85  & 0.82  & 0.23  & 0.05  & 0.04  & 0.68 \\
  \textit{North American Plant B}  & 0.95  &  0.9 & 0.22  & 0.02  & 0.00  & 0.75 \\
   \textit{North American Plant C} & 0.95 & 0.81 & 0.15  & 0.04 & 0.01 & 0.8 \\
   \textit{North American Plant D} & 0.95 & 0.86 & 0.21 & 0.03 & 0.01 & 0.75  \\
   \textit{North American Plant E} & 0.85  & 0.81 & 0.25 & 0.06 & 0.04 & 065 \\
 \hline
\end{tabular}
}
\caption{Quality Assurance (QA) performance of the model for five selected plants in North America. Metrics were calculated using the definitions of Adapted True Positive, False Negative, False Positive, and True Negative as previously described. For brevity, 'Adapted' is omitted in the column names.}
\label{tab:performance}
\end{table}

In this section, we detail the approach for evaluating the proposed model at Ford to ensure its quality. The trained model estimates the time until a shortfall event occurs related to the shipment of vehicle parts between a supplier and a Ford plant. To assess the model's performance, we modify and adapt the standard classification metrics to fit our specific setting.

In binary classification, the two standard metrics for evaluation are Precision and Recall (see \cite{buckland1994}). Binary classification involves two labels: positive and negative. For a given classifier, True Positives (TP) are the instances where the classifier correctly identifies positives. True Negatives (TN) are the instances where the classifier correctly identifies negatives. False Positives are instances where the classifier incorrectly predicts positives when the true labels are negative. Conversely, False Negatives (FN) are instances where the classifier incorrectly predicts negatives when the true labels are positive. Formally, Precision and Recall are defined as,

\[
\begin{aligned}
&\text{Precision} = \cfrac{\text{TP}}{\text{TP} + \text{FP}} \\
&\text{Recall} = \cfrac{\text{TP}}{\text{TP} + \text{FN}}.
\end{aligned}
\]

Precision and recall range from $0$ to $1$. The closer these values are to $1$, the better the classifier's performance. Specifically, recall quantifies the proportion of actual positive data points that the classifier successfully identifies. Precision, on the other hand, measures the proportion of predicted positive cases that are, in fact, positive. Informally, recall measures how effectively a classifier identifies positive events (shortfalls in this context), while precision measures the frequency with which the classifier avoids generating false alarms. These two metrics offer a comprehensive view of a model's classification performance. 

Given that the model proposed in Section \ref{model} estimates the time until a shortfall, we first define a forecasting horizon, $\Delta$ (in days), and a margin-of-error, $\epsilon$ (in days), to evaluate its performance. We introduce the following definitions to adapt the concepts of True Positives, False Positives, True Negatives, and False Negatives to our specific problem. Given an observation time $t_0$, and the forecasting window $\Delta$ and margin-of-error $\epsilon$, $t_0 + \Delta$ represents a time point $\Delta$ days from $t_0$. Adapted True Positives (ATP) are defined as the number of shortfalls that occur exactly $\Delta$ days from $t_0$ and are estimated by the model within $\epsilon$ days of $t_0 + \Delta$. The difference between the model's estimated time and $t_0 + \Delta$ must fall within the range $[- \epsilon, \epsilon]$ days. Adapted False Positives (AFP) are instances where the model predicts a shortfall time exactly $\Delta$ days from $t_0$, but no shortfalls actually occur within $\epsilon$ days of $t_0 + \Delta$.

Conversely, Adapted True Negatives (ATN) are instances where the model estimates the shortfall time to be more than $\Delta$ days from $t_0$, and no disruptions occur between $t_0$ and $t_0 + \Delta$. Adapted False Negatives (AFN)  are instances where the model predicts a shortfall time beyond $\Delta$ days from $t_0$, while shortfalls actually occur within $\Delta$ days from $t_0$.

With these definitions, Adapted Recall is computed as the ratio of Adapted True Positives (ATP) to the total of ATP and Adapted False Negatives (AFN). Likewise, Adapted Precision is the proportion of ATP relative to the sum of ATP and Adapted False Positives (AFP). Table \ref{tab:performance} shows the metric performance for our proposed model. Additionally, we include the Normalized Confusion Matrix in Table \ref{tab:performance}. The Confusion Matrix offers a detailed breakdown of the four cases: True Positives, False Negatives, False Positives, and True Negatives. The normalization is done by dividing each case by the total number of cases.

\begin{table}[!ht]
\centering
\begin{tabular}{|c||c|c|} 
\hline
\multirow{2}{*}{\textbf{Model on Data}} & \multicolumn{2}{|c|}{\textbf{Performance Metrics}}  \\
\cline{2-3}
 & Precision & Recall  \\
 \hline
  \textit{Seq2Surv on North American Plant A's Data} & 0.62 & 0.3 \\
  \textit{Seq2Surv on North American Plant E's Data}  & 0.52  &  0.21 \\
 \hline
   \textit{Heterogeneous Seq2Surv on North American Plant A's Data} & 0.85 & 0.82  \\
   \textit{Heterogeneous Seq2Surv on North American Plan E's Data} & 0.85 & 0.81 \\
 \hline
\end{tabular}
\caption{Model Performance Comparison of Seq2Surv and Heterogeneous Seq2Surv on North American Plant A and B.  Heterogeneity presents a critical challenge for Seq2Surv.}
\label{tab:comp}
\end{table}

We conducted 20 iterations of Quality Assurance (QA) testing from August 1, 2023, to January 13, 2024 for the result in Table \ref{tab:performance}. For each observation week, we trained the model on the cumulative data up to that week and used it to predict disruptions in the near future. We computed Adapted Precision, Adapted Recall, and the Normalized Confusion Matrix using the above definitions for Adapted True Positives, True Negatives, False Positives, and False Negatives, with parameters forecasting horizon $\Delta = 28$ days and margin-of-error $\epsilon = 7$ days. The metrics in Table \ref{tab:performance} are averaged over the 20 iterations. The Quality Assurance (QA) evaluation was conducted at five selected Ford manufacturing facilities in North America.

The metrics provided are the averages obtained from 20 iterations of QA testing. The model demonstrated on average precision exceeding $0.85$ and a recall surpassing $0.8$ across the five plants during the 20-week QA period. Additionally, the normalized confusion matrix indicates that the combined error rate of false positive and false negative errors was below $10$\% on average. Furthermore, we present performance comparison between Seq2Surv, as described in \cite{li2022}, and our Heterogeneous Seq2Surv proposed in Section \ref{model}, as shown in Table \ref{tab:comp}. The heterogeneity in plant, supplier, and part groups poses significant challenges for Seq2Surv. However, by incorporating heterogeneous behaviors in the data into the model, our Heterogeneous Seq2Surv demonstrates substantially improved performance. For a comparison between Seq2Surv and other models, please refer to \cite{li2022}.

\subsection{Explainable AI with SHAP}

\begin{figure}[ht!]
    \centering
    \includegraphics[scale=0.55]{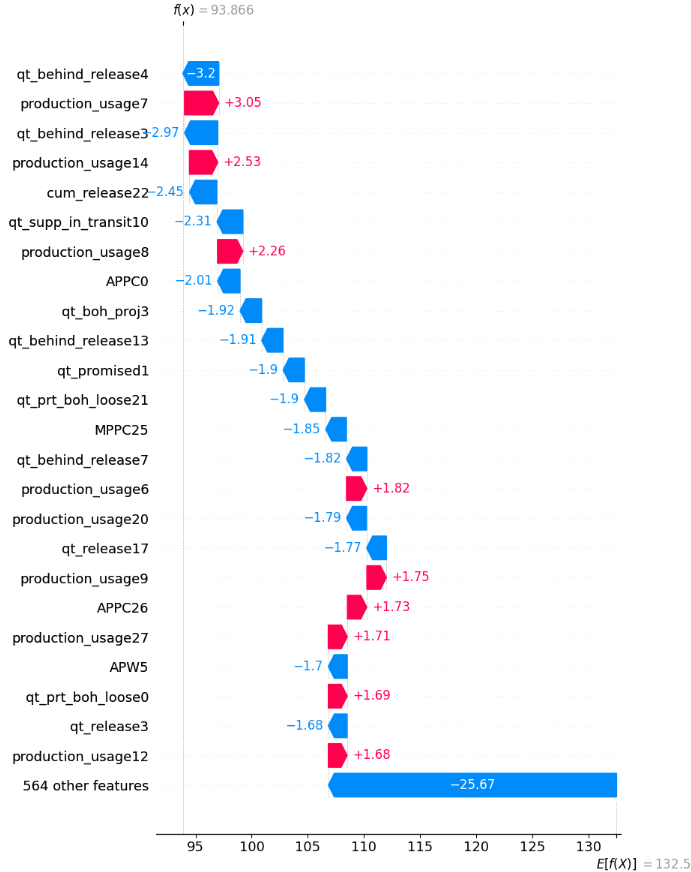}
    \caption{Utilizing the SHAP (SHapley Additive exPlanations) framework to calculate feature importance for a specific prediction instance made by the trained model.}
    \label{fig:XAI}
\end{figure}

The proposed model demonstrates desirable predictive performance, largely due to the Sequence-to-Sequence Neural Network's capability to capture complex non-linear relationships. However, despite their effectiveness, Deep Learning models often face criticism for their "black-box" nature. The complexity of these models frequently makes it difficult to understand their underlying mechanisms or interpret internal embeddings and representations (see \cite{wang2018}). This lack of transparency is a significant challenge in a multi-departmental corporation like Ford, where it is essential to adequately communicate the model's behaviors and decision-making processes to provide actionable insights to the Business department.

Explainable Artificial Intelligence (XAI) is an emerging field dedicated to developing principles, frameworks, and tools to elucidate the reasoning behind AI decisions and predictions (see \cite{phillips2021}). A prominent technique in this area is the Shapley value, which is based on cooperative game theory (see \cite{lundberg2017}).

Introduced by Lloyd Shapley in 1951, the Shapley value is a technique for fairly allocating profits among players in a cooperative game, adhering to four fairness axioms (see \cite{roth1988shapley}). In machine learning, Shapley values can be applied by treating features as players and the model's behavior as the profit. This approach allows us to quantify the impact of each feature. The SHAP (SHapley Additive exPlanations) framework, popularized by \cite{lundberg2017}, utilizes Shapley values to explain individual predictions. Figure \ref{fig:XAI} demonstrates the application of the SHAP framework to a specific prediction instance using the model described in Section \ref{model}. The waterfall plot illustrates the important factors influencing the model's predicted time-to-disruption, such as the progression of `'qt\_behind\_release`' over 13, 7, 4, and 3 days ago, and the progression of `'production\_usage`' over 27, 14, 12, 9, 8, 7, and 6 days ago, among others. This information enables a deeper investigation into the data, helping to uncover the potential reasons behind the model's prediction for this specific instance.

\section{Conclusion}

In this paper, we first outline the process of constructing a dataset of multivariate time series for forecasting first-tier supply chain disruptions. We carefully selected features representing key aspects of classical Factory Physics, such as capacity, variability, utilization, and processing time.

The dataset, comprising over five hundred thousand individual time series, presents significant technical challenges due to its complexity and scale. Although these time series exhibit some commonalities, they also demonstrate substantial heterogeneity within specific subgroups, making traditional industrial and statistical models inadequate. To address these challenges, we propose a novel methodology that integrates an enhanced Attention Sequence-to-Sequence Deep Learning architecture with Neural Network Embeddings to model group effects, combined with a Survival Analysis model. This approach is designed to effectively capture complex heterogeneous data patterns related to operational disruptions. Our model has achieved strong performance, with at least $85$\% precision and $80$\% recall during the Quality Assurance (QA) phase across Ford's five North American plants.

To mitigate the criticism of Deep Learning models as ``black boxes,'' we employed the SHAP (SHapley Additive exPlanations) framework to elucidate feature importance in the model’s predictions. This technique provides valuable insights that can guide actionable strategies for business partners. Our work highlights the potential of advanced machine learning techniques to manage and mitigate supply chain risks effectively within the automotive industry.

\section*{Acknowledgement}

I am grateful to Zhen Jia for her support in cleaning and integrating multiple raw data sources into a unified dataset, as well as for providing clear and detailed explanations of field names and their meanings. I would also like to thank Oleg Gusikhin for his thoughtful review of our work and constructive feedback that enhanced the quality of this study.

\newpage

\printbibliography

\end{document}